# Color Space Transformation Network

Alexandros Karargyris
National Library of Medicine, Maryland, USA
alexandros.karargyris@nih.gov

**Abstract**
Deep networks have become very popular over the past few years. The main reason for this widespread use is their excellent ability to learn and predict knowledge in a very easy and efficient way. Convolutional neural networks and auto-encoders have become the normal in the area of imaging and computer vision achieving unprecedented accuracy levels in many applications. The most common strategy is to build and train networks with many layers by tuning their hyper-parameters. While this approach has proven to be a successful way to build robust deep learning schemes it suffers from high complexity. In this paper we introduce a module that learns color space transformations within a network. Given a large dataset of colored images the color space transformation module tries to learn color space transformations that increase overall classification accuracy. This module has shown to increase overall accuracy for the same network design and to achieve faster convergence for the same number of epochs. It is part of a broader family of image transformations (e.g. spatial transformer network [4]).

## 1. Introduction

Convolutional neural networks (CNNs) have become very successful in computer vision applications over recent years. With GPUs becoming more of a commodity for research purposes and better techniques, it has become possible to train very large networks and achieve higher performance for complex computer vision problems. Intermediate max-pooling layers contribute to better performance. In 2012, Dan Ciresan et al. significantly improved upon the best performance in the literature for multiple image databases, including the MNIST database, the NORB database, the HWDB1.0 dataset (Chinese characters), the CIFAR10 dataset [1] and the ImageNet dataset [2]. [3]

Since then there have been efforts to improve overall performance by: a) increasing the depth of the networks, b) combining different network architectures (LSTM, CNN, etc.), c) modifying layer cells to explicitly optimize a function (i.e. spatial transformation [4]). Our works falls into the last category.

## 2. Color Space Transformation Module

In this paper we introduce a very simple idea: let the neural network identify the best color space transformations for a given colored dataset. Figure 1 shows how the module is used within a network.

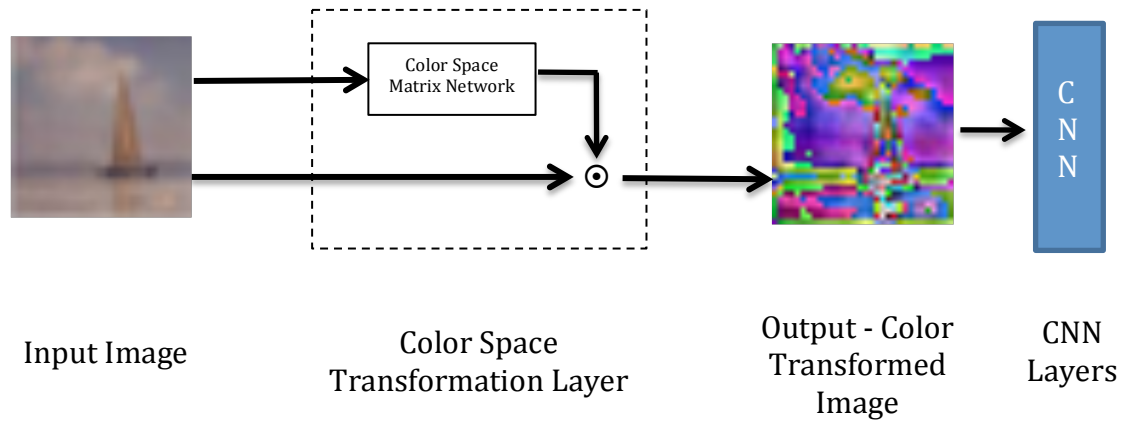

Figure 1. Architecture of neural network with color space transformation layer

**Figure 1. Architecture of neural network with color space transformation layer**

As seen in Figure 1 the network is trained with a dataset of colored images as usual. Each image's pixels serve as an input to the color space matrix transformer where they are sampled to generate a 3x3 transformation matrix (**W**). The matrix **W** is then applied on the input image pixels to map them to another space. Subsequently, this transformed image is fed as input to conventional CNN layers. Because of the linearity of the transformation the backpropagation of the loss can be applied.

Basically after each epoch the Color Space Matrix Network predicts the best transformation (parameters in W) to be applied on an input image to increase the network's classification accuracy.

## 3. Experiments

We used the CIFAR10 dataset to test the performance of the proposed module. For this we compared against a baseline CNN (3 layers deep). We used the same CNN and we added the color space transformation layer as shown in Figure 1. Figure 2 shows the output of the color space transformation layer while Figure 3 shows the classification error for both networks.

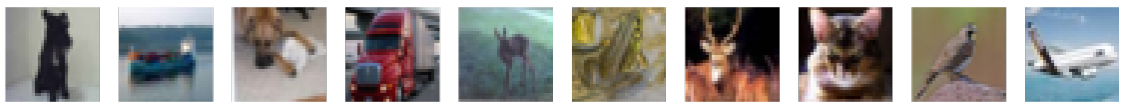

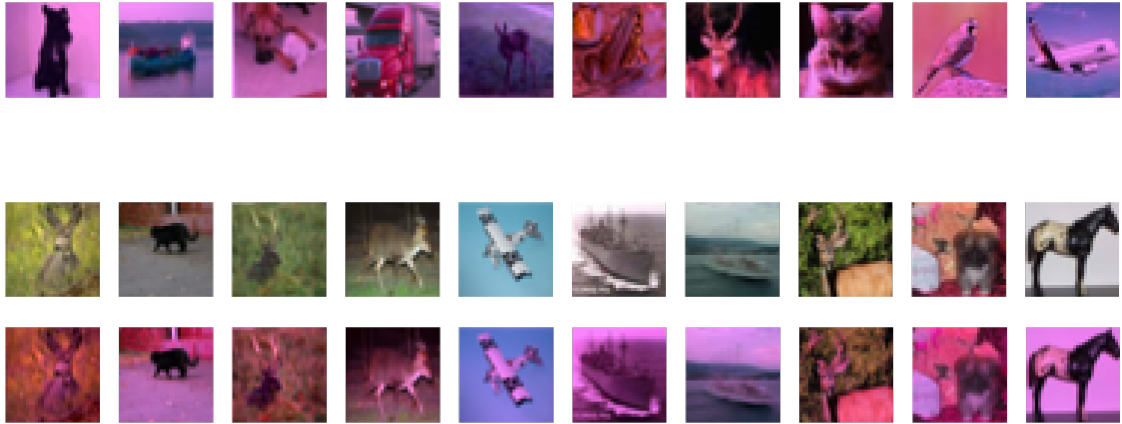

**Figure 2. Some example outputs of the color space transformation module from CIFAR10 after training them on CNN coupled with proposed color space transformation module. Notice that the color space transformation tends to be global.**

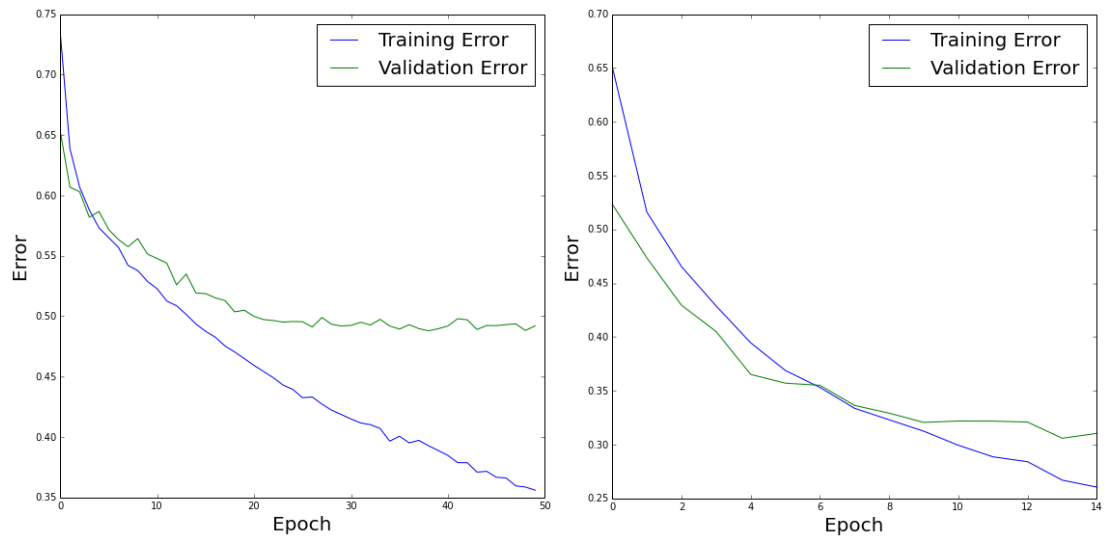

**Figure 3. Classification Error. Left: Baseline CNN, Right: CNN + Color Space Transformation Network**

Not only does the proposed architecture with the new module converges faster (e.g. 8 epochs vs. 15) but it also reaches higher accuracy (~68% vs. ~50%).

The source code for the network and the experiments is written in Lasagne and it is available here: Lasagne (http://lasagne.readthedocs.org/) is an excellent research tool to build and run neural networks in a quick and efficient way.

## 4. Discussion & Future Work

This work proposed a simple approach to improve classification accuracy in colored images using a neural network. It consists of a module (layer) that optimizes its weights to create a color transformation that increases discrimination power between color channels and as a result increases the network's overall accuracy. Its main advantage is its simplicity to plug the layer into any network.

While performance for this module was tested and compared in a popular set (i.e. CIFAR10) more experiments need to be performed on other datasets to validate its efficiency. Additional comparisons need to be made against other methodologies that are used to create color-space transformations (i.e. Karhunen-Loeve color transform). Space reduction methods such as PCA and Whitening are used as pre-processing step in images to increase discrimination power. Therefore it is important from a theoretical point of to identify any similarity.